\documentclass[twoside]{article}
\usepackage{ecj,palatino,epsfig,latexsym,natbib}
\usepackage{dsfont,xspace}
\usepackage[vlined,linesnumbered,ruled,titlenotnumbered]{algorithm2e}
\usepackage[noend]{algpseudocode}
\usepackage{multirow}
\usepackage{dsfont}
\usepackage{amsmath}
\usepackage{amssymb}
\usepackage{xcolor}
\usepackage{setspace}

\usepackage{graphicx}
\usepackage{amsmath}
\usepackage{amssymb}
\usepackage{amsthm}

\usepackage{fixltx2e}
\newcommand{\ignore}[1]{}
\newcommand{\remark}[1]{\textcolor{red}{#1}}

\newcommand{\dist}{d}
\newcommand{\topt}{$2$-OPT\xspace}
\newcommand{\EAd}{($\mu$+$\lambda$)-EA\textsubscript{D}\xspace}

\newcommand{\ie}{i.\,e.\xspace}

\begin{document}
\singlespacing
\ecjHeader{x}{x}{xxx-xxx}{2020}{Feature-based Diversity Optimization for Problem Instance Classification}{W. Gao, S. Nallaperuma, F. Neumann}
\title{\bf Feature-based Diversity Optimization for Problem Instance Classification}  

\author{\name{\bf Wanru Gao} \hfill \addr{kelly\_gwr@hotmail.com}\\ 
        \addr{School of Information Engineering, Zhengzhou University, 
        Zhengzhou, Henan, China}
\AND
       \name{\bf Samadhi Nallaperuma} \hfill \addr{snn26@cam.ac.uk}\\
        \addr{University of Cambridge, 
        Cambridge, UK}
 \AND
       \name{\bf Frank Neumann} \hfill \addr{frank.neumann@adelaide.edu.au}\\
        \addr{School of Computer Science, University of Adelaide, 
        Adelaide, Australia}
}

\maketitle

\begin{abstract}

Understanding the behaviour of heuristic search methods is a challenge. This even holds for simple local search methods such as $2$-OPT for the Traveling Salesperson problem. In this paper, we present a general framework that is able to construct a diverse set of instances which are hard or easy for a given search heuristic. Such a diverse set is obtained by using an evolutionary algorithm for constructing hard or easy instances which are diverse with respect to different features of the underlying problem. Examining the constructed instance sets, we show that many combinations of two or three features give a good classification of the TSP instances in terms of whether they are hard to be solved by $2$-OPT. 

\end{abstract}


\section{Introduction}
Heuristic search methods such as local search, simulated annealing, evolutionary algorithms and ant colony optimization have been shown to be very successful for various combinatorial optimization problems. Although they usually don't come with performance guarantees on their runtime and/or approximation behaviour, they often perform very well in several situations. Understanding the conditions under which optimization algorithms perform well is essential for automatic algorithm selection, configuration and effective algorithm design. In both the artificial intelligence (AI)~\citep{Xu2008,Hutter2014,Vilalta02,EggenspergerHHL15,FeurerSH15} and operational research~\citep{SML12,vanHemert2006} communities, this topic has become a major point of interest.

The feature-based analysis of heuristic search algorithms has become an important part in understanding such type of algorithms~\citep{AMAI/Mersmann2013,SmithMiles2010}. This approach characterizes algorithms and their performance for a given problem based on features of problem instances. Thereby, it provides an important tool for bridging the gap between pure experimental investigations and mathematical methods for analysing the performance of search algorithms~\citep{Neumann2010,KotzingNRW12,EnglertRV14}.
Current methods for the feature-based analysis are based on constructing hard and easy instances for an investigated search heuristic and a given optimization problem by evolving instances using an evolutionary algorithm~\citep{AMAI/Mersmann2013,Nallaperuma0NBMT13,Nallaperuma0N14}. This evolutionary algorithm constructs problem instances where the examined algorithm either shows a bad (good) approximation behaviour and/or requires a large (small) computational effort to come up with good or optimal solutions. Although the evolutionary algorithm for constructing such instances is usually run several times to obtain a large set of hard (easy) instances, the question arises whether the results in terms of features of the instances obtained give a good characterization of problem difficulty.

In the paper, we propose a new approach of constructing hard and easy instances. Following some recent work on using evolutionary algorithms for generating diverse sets of instances that are all of high quality~\citep{ppsn/UlrichBT10,gecco/UlrichBZ10}, we introduce an evolutionary algorithm which maximizes diversity of the obtained instances in terms of a given feature. The recent work of ~\cite{DBLP:journals/cor/Smith-MilesB15} also considers diversity aspect of the instances. However, our notion of diversity is fundamentally different from this work. Our notion of diversity follows from the evolutionary diversity optimization mechanisms where as theirs considers a notion of the distance from manually specified target points. Our approach allows to generate a set of instances that is much more diverse with respect to the problem feature at hand. Carrying out this process for several features of the considered problem and algorithm gives a much better classification of instances according to their difficulty of being solved by the considered algorithm.

To show the benefit of our approach compared to previous methods, we consider the classical \topt algorithm for the TSP. Previous feature-based analyses have already considered hard and easy instances in terms of approximation ratio and analyzed the features of such hard (easy) instances obtained by an evolutionary algorithm.
The experimental results of our new approach show that diversity optimization of the features results in an improved coverage of the feature space over classical instance generation methods. In particular, the results show that for some combinations of two features, it is possible to classify hard and easy instances into two clusters with a wider coverage of the feature space compared to the classical methods. Moreover, the three-feature combinations further improve the classification of hard and easy instances for most of the feature combinations. Furthermore, a classification model is built using these diverse instances that can classify TSP instances based on hardness for \topt. 

This paper is extended from its conference version which was published in PPSN 2016~\citep{GaoNN16}. The approach proposed in the conference paper has been applied in artistic image variants analysis~\citep{DBLP:conf/gecco/AlexanderKN17} and examined with different diversity measurement such as discrepancy~\citep{DBLP:conf/gecco/NeumannGDN018} and popular indicators from the area of evolutionary multi-objective optimisation~\citep{DBLP:conf/gecco/NeumannG0019}. 
Our further research focuses on a weighted version of feature-based population diversity measurement and involving problem hardness as an extra feature value, which will be detailed in Section~\ref{sec:weighted} and~\ref{sec:divAP}.

The remainder of this paper is organized as follows. Firstly, we introduce the Euclidean TSP and the background on feature based analysis in Section~\ref{sec:background}. Afterwards, we state our diversity optimization approach for evolving instances according to feature values in Section~\ref{sec:approach} and report on the impact of diversity optimization in terms of the range of feature values in Section~\ref{sec:range}. As feature values can be very diverse both for easy and hard instances, in Section~\ref{sec:exp} we consider the combinations of several features for instance classification afterwards. We then build a classification model that can classify instances based on hardness and the results are included in Section~\ref{sec:svm}. Section~\ref{sec:divAP} discusses how to maximize the diversity over problem hardness and the classification. The paper is finalized with some conclusions in Section~\ref{sec:conclusions}.

\section{Background}
\label{sec:background}
We consider the classical NP-hard Euclidean Traveling Salesperson problem (TSP) as the example problem for evolving hard and easy instances which have a diverse set of features. Our methodology can be applied to any optimization problem, but using the TSP in our study has the advantage that it has already been investigated extensively from different perspectives including the area of feature-based analysis.

The input of the problem is given by a set $V = \{v_1 , . . . , v_n \}$ of $n$ cities in the Euclidean plane and Euclidean distances $\dist : V \times V \rightarrow \mathds{R}_{\geq 0}$ between the cities. The goal is to find a Hamiltonian cycle whose sum of distances is minimal.
A candidate solution for the TSP is often represented by a permutation $\pi= (\pi_1, \ldots, \pi_n$) of the $n$ cities
and the goal is to find a permutation $\pi^*$ which minimizes the tour length given by
\[
  \label{eq:fitness}
  c(\pi) = d(\pi_n,\pi_1) + \sum_{i=1}^{n-1} d(\pi_i, \pi_{i+1}).
\]

For our investigations cities are always in the normalized plane $[0,1]^2$, \ie each city has an $x$- and $y$-coordinate in the interval $[0,1]$.
In following, a TSP instance always consists of a set of $n$ points in $[0,1]^2$ and the Euclidean distances between them.

Local search heuristics have been shown to be very successful when dealing with the TSP and the most prominent local search operator is the \topt operator~\citep{Croes1958}. The resulting local search algorithm starts with a random permutation of the cities and repeatedly checks whether removing two edges and reconnecting the two resulting paths by two other edges leads to a shorter tour. If no improvement can be found by carrying out any \topt operation, the tour is called locally optimal and the algorithm terminates. 

The key factor in the area of feature-based analysis is to identify the problem features and their contribution to the problem hardness for a particular algorithm and problem combination. This can be achieved through investigating hard and easy instances of the problem.
\ignore{
The conventional approach to generate hard or easy instances was to set the values of a specified set of features in order to modify the problem difficulty level. \remark{Frank: I don't understand this sentence, paragraph is much too long, no need for extensive literature review}
Then the algorithm performance is measured on these instances. This approach was later criticized due to  the difficulty of generating diverse random instances and the restrictedness of randomly generated benchmark datasets in the spectrum of difficulty~\citep{SmithMiles2010}.

Van Hemert~\citep{vanHemert2006} has proposed an approach that is based on an evolutionary algorithm that evolves instances based on the performance of the investigated algorithm. After this study on the Lin-Kernighan algorithm~\citep{LinK1973}, there have been several other studies following this approach for instance generation ~\citep{vanHemert2006,SML12,AMAI/Mersmann2013,SmithMiles2010}.
} 
Using an evolutionary algorithm, it is possible to evolve sets of hard and easy instances by maximizing or minimizing the fitness (tour length in the case of the TSP) of each instance~\citep{vanHemert2006,SML12,AMAI/Mersmann2013,SmithMiles2010}. However, none of these approaches have considered the diversity of the instances explicitly. Within this study we expect to improve the evolutionary algorithm based instance generation approach by introducing diversity optimization.

The structural features are dependent on the underlying problem. In \cite{AMAI/Mersmann2013}, there are 47 features in different feature classes used to provide an understanding of algorithm performance for the TSP. The different feature classes established are distance features, mode features, cluster features, centroid features, MST features, angle features and convex hull features. The feature values are regarded as indicators which allow to predict the performance of a given algorithm on a given instance. 

\section{Feature-Based Diversity Optimization}
\label{sec:approach}

\begin{algorithm}[t]
{
   	Initialize the population $P$ with $\mu$ TSP instances of approximation ratio at least $\alpha_h$.\\
	Let $C \subseteq P$ where $|C| = \lambda$.\\
	For each $ I \in C$, produce an offspring $I'$ of $I$ by mutation. If $\alpha_A(I')\geqslant\alpha_{h}$, add $I'$ to $P$. \\
	While $|P| > \mu$, remove an individual $I=\arg \min_{J \in P} d(J,P)$ uniformly at random.\\ 
	
 	Repeat step 2 to 4 until termination criterion is reached.\\
} 
 \caption{$(\mu+\lambda)$-$EA_{D}$}
\label{EA}
\end{algorithm}

In this section, we introduce our approach of evolving a diverse set of easy or hard instances which are diverse with respect to important problem features.
As in previous studies, we measure hardness of a given instance by the ratio of the solution quality obtained by the considered algorithm and the value of an optimal solution.

The approximation ratio of an algorithm $A$ for a given instance $I$ is
defined as
 \[
  \alpha_A(I) = A(I) / OPT(I)
\]
where $A(I)$ is the value of the solution produced by algorithm $A$ for the given instance $I$, and $OPT(I)$ is value of an optimal solution for instance $I$. Within this study, $A(I)$ is the tour length obtained by \topt for a given TSP instance $I$ and $OPT(I)$ is the optimal tour length which we obtain in our experiments by using the exact TSP solver Concorde~\citep{Applegate02}.

We propose to use an evolutionary algorithm to construct sets of instances of the TSP that are quantified as either easy or hard in terms of approximation and are diverse with respect to underlying features of the produced problem instances.
Our evolutionary algorithm (shown in Algorithm~\ref{EA}) evolves instances which are diverse with respect to given features and meet given approximation ratio thresholds. 

The algorithm is initialized with a population $P$ consisting of $\mu$ TSP instances which have an approximation ratio at least $\alpha_h$ in the case of generating a diverse set of hard instances. In the case of easy instances, we start with a population where all instances have an approximation ratio of at most $\alpha_e$ and only instances of approximation ratio at most $\alpha_e$ can be accepted for the next iteration. In each iteration, $\lambda \leq \mu$ offspring are produced by selecting $\lambda$ parents and applying mutation to the selected individuals. Offsprings that don't meet the approximation threshold are rejected immediately.

The new parent population is formed by reducing the set consisting of parents and offsprings satisfying the approximation threshold until a set of $\mu$ solutions is achieved. This is done by removing instances one by one based on their contribution to the diversity according to the considered feature.

The core of our algorithm is the selection among individuals meeting the threshold values for the approximation quality according to feature values. Let $I_1, \ldots, I_k$ be the elements of $P$  and $f(I_i)$ be their feature values. Furthermore, assume that $f(I_i) \in [0,R]$, i.e. feature values are non-negative and upper bounded by $R$.

We assume that $f(I_1) \leq f(I_2) \leq \ldots \leq f(I_k)$ holds. The feature-based diversity contribution of an instance $I$ to a population of instances $P$ is denoted as $d(I, P)$ which is based on other individuals in the population.


Let $I_i$ be an individual for which $f(I_i)\not = f(I_1)$ and $f(I_i)\not = f(I_k)$. We set

\[
d(I_i,P) =(f(I_i) - f(I_{i-1})) \cdot (f(I_{i+1}) - f(I_i))
\]
which assigns the diversity contribution of an individual based on the next smaller and next larger feature values.
If $f(I_i)  = f(I_1)$ or $f(I_i) = f(I_k)$, we set $d(I_i,P)=R^2$ if there is no other individual $I\not=I_i$ in $P$ with $f(I)= f(I_i)$ and $d(I_i,P)=0$ otherwise.
This implies an individual $I_i$ with feature value equal to any other instances in the population gains $d(I_i,P) = 0$.

Furthermore, an individual with the unique smallest and largest feature value always stays in the population when working with $\mu \geq 2$.

\ignore{
\[
c'(I_j,P') =(f(I_j) - f(I_{j-1})) \cdot (f(I_{j+1}) - f(I_j))
\]
 set $c(I, P)=  c'(I_j,P')$
where $I_j$ is the instance equal to $I$.

where $c(I)$ is a contribution value to diversity and $p$ is a penalty in the case that there is an instance in the population with the same 
}
\ignore{
\begin{algorithm}
{
Input: Instance $s$, standard deviation of the normal distribution $\sigma$.

	Select a city ${x,y} \in s$ uniformly at random.\\
	Let $x' := N(x,\sigma^2)$ and $y' := N(y,\sigma^2)$.\\
	If $x' \in (0,1)$, let $x:=x'$.  If $y' \in (0,1)$, let $y:=y'$.\\  
}
\caption{Normal Mutator}
\label{normalMutation}
\end{algorithm}
}
\label{features}

In \cite{AMAI/Mersmann2013},  $47$ features of TSP instances for characterizing easy and hard TSP instances have been studied.
We consider $7$ features coming from different feature classes which have shown to be well suited for classification and prediction. These features are: 
\begin{itemize}
	\item \emph{angle\_mean} :- mean value of the angles made by each point with its two nearest neighbor points
	\item \emph{centroid\_mean\_distance\_to\_centroid} :- mean value of the distances from the points to the centroid
	\item \emph{chull\_area} : - area covered by the convex hull
	\item \emph{cluster\_10pct\_mean\_distance\_to\_centroid} :- mean value of the distances to cluster centroids at $10\%$ levels of reachability 
	\item \emph{mst\_depth\_mean} :- mean depth of the minimum spanning tree
	\item \emph{nnds\_mean} :- mean distance between nearest neighbours
	\item \emph{mst\_dists\_mean} :- mean distance of the minimum spanning tree
\end{itemize}

We refer the reader to \cite{AMAI/Mersmann2013} for a detailed explanation for each feature.
We carry out our diversity optimization approach for these features and use the evolutionary algorithm to evolve for each feature a diverse population of instances that meets the approximation criteria for hard/easy instances given by the approximation ratio thresholds.

All programs in our experiments are written in R and run in R environment~\citep{rManual}. We use the functions in tspmeta package~\citep{AMAI/Mersmann2013} to compute the feature values.

The setting of the evolutionary algorithm for diversity optimization used in our experiments is as follows.
 We use $\mu=30$ and $\lambda=5$ for the parent and offspring population size, respectively. 
The \topt algorithm is executed on each instance $I$ five times with different initial solutions and we set $A(I)$ to the average tour length obtained. The examined instance sizes $n$ are $25$, $50$ and $100$, which are denoted by the number of cities in one instance.
 Based on previous investigations in \cite{AMAI/Mersmann2013} and initial experimental investigations, we set $\alpha_e=1$ for instances of size $25$ and $50$, and $\alpha_e=1.03$ for instances of size $100$. Evolving hard instances, we use  $\alpha_h=1.15, 1.18, 1.2$ for instances of size $n=25,50,100$, respectively.
The mutation operator picks in each step one city for the given parent uniformly at random and changes its $x$- and $y$-coordinator by choosing an offset according to the Normal-distribution with standard deviation $\sigma$. Coordinates that are out of the interval are reset to the value of the parent. Based on initial experiments we use two mutation operators with different values of $\sigma$. We use $\sigma=0.025$ with probability $0.9$ and $\sigma=0.05$ with probability $0.1$ in a mutation step.
The evolutionary algorithm terminates after $10,000$ generations which allows to obtain a good diversity for the considered features. 
For each $n=25,50,100$ and each of the $7$ features, a set of easy and hard instances are generated, which results in $42$ independent runs of the \EAd.


\begin{figure}
\centering

\includegraphics[height=4.9cm]{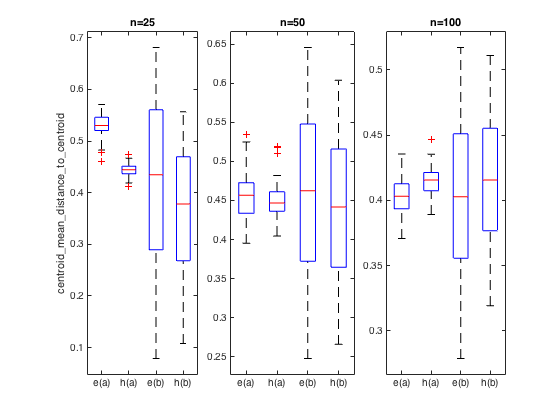}
 \includegraphics[height=4.9cm]{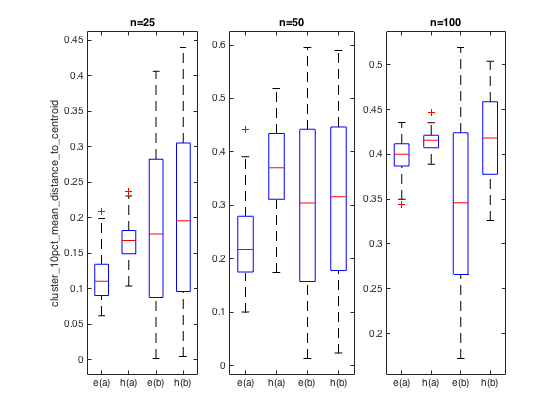}
\caption{\small{(left)The boxplots show the distribution of centroid\_mean\_distance\_to\_centroid feature values of a population consisting of $100$ different hard or easy TSP instances of different number of cities without or with diversity mechnism.(right)The boxplots show the distribution of cluster\_$10\%$\_mean\_distance\_to\_centroid  feature values of a population consisting of $100$ different hard or easy TSP instances of different number of cities without or with diversity mechnism. Easy and hard instances from conventional approach and diversity optimization are indicated by e(a), h(a) and e(b), h(b) respectively. }}
\label{fig:feature-range}
\end{figure}


\section{Range of Feature Values}
\label{sec:range}

We first evaluate our diversity optimization approach in terms of the diversity that is obtained with respect to a single feature. Focusing on a single feature in each run provides the insight of the possible range of a certain feature value for hard or easy instances.  The previous study~\citep{AMAI/Mersmann2013}, suggests that there are some differences in the possible range of feature values for easy and hard instances. We study the effect of the diversity optimization on the range of features by comparing the instances generated by diversity optimization to the instances generated by the conventional approach in~\citep{AMAI/Mersmann2013}. Evolving hard instances based on the conventional evolutionary algorithm, the obtained instances have mean approximation ratios of $1.12$ for $n=25$, $1.16$ for $n=50$, and $1.18$ for $n=100$. For easy instances, the mean approximation ratios are $1$ for  $n=25,50$ and $1.03$ for $n=100$.

Figure~\ref{fig:feature-range} (left) presents the variation of the mean distance of the distances between points and the centroid feature (\emph{centroid\_mean\_distance\_to\_centroid}) for hard and easy instances of the three considered sizes $25$, $50$ and $100$. Each set consists of 100 instances  generated by independent runs~\citep{AMAI/Mersmann2013}. As shown in Figure~\ref{fig:feature-range} (left), the feature values of easy and hard instances are in different ranges. For example, for the hard instances of size $100$ the median value (indicated by the red line) is $0.4157$ while it is only $0.4032$ for the easy instances. The respective range of the feature value is $0.0577$ for the hard instances and $0.0645$ for the easy instances. For the instances generated by diversity optimization (easy and hard instances are indicated by e (b) and h (b) respectively),  there is a difference in the median feature values for the hard and easy instances similar to the instances generated by the conventional approach. Additionally, the range of the feature values for both the hard and easy instances has significantly increased. For example, for the instance size $100$, the median value for easy instances is $0.4028$ and the range is $0.2382$. For the hard instances of the same size, the median is $0.04157$ while the range is $0.1917$ (see Figure~\ref{fig:feature-range} (left)). 

Similarly, Figure~\ref{fig:feature-range} (right) presents the variation of cluster $10\%$ distance to centroid (\emph{cluster\_10pct\_distance\_to\_centroid}) feature for the hard and easy instances generated by the conventional approach (indicated by (e (a) and h (a)) and for the hard and easy instances generated by diversity optimization (indicated by (e (b) and h (b))). The general observations from these box plots are quite similar to the observations from the \emph{mst\_dist\_mean} shown in Figure~\ref{fig:feature-range} (left). For the easy instances of size $100$, the range of the feature value is $0.0919$ for conventional instances and $0.3471$ for the instances generated by diversity optimization. Similarly, for the hard instances the range of the feature values has increased from $0.0577$ to $0.1776$ by the diversity optimization approach. As shown in Figure~\ref{fig:feature-range} (right), there is a significant increase in the range for other instance sizes as well. Improved ranges in feature values are observed for all considered feature. \ignore{however, due to space limitations these are not included in the paper.}

\begin{table}[]
\resizebox{\textwidth}{!}{%
\begin{tabular}{|cllllllllllllll|}
\hline
\multicolumn{1}{|l}{} &  &  & \multicolumn{4}{|l|}{centroid\_mean\_distance\_to\_centroid} & \multicolumn{4}{l|}{distance\_mean} & \multicolumn{4}{l|}{nnd\_mean} \\ \hline
\multicolumn{1}{|l|}{n} &  & \multicolumn{1}{|l|}{algorithm} & min & max & avg & \multicolumn{1}{l|}{stdev} & min & max & avg & \multicolumn{1}{l|}{std} & min & max & avg & std \\ \hline
\multicolumn{1}{|c|}{\multirow{4}{*}{25}} & \multicolumn{1}{l|}{\multirow{2}{*}{easy}} & \multicolumn{1}{l|}{conventional} & 0.4603 & 0.5708 & 0.5303 & \multicolumn{1}{l|}{0.0214} & 0.6125 & 0.7613 & 0.7078 & \multicolumn{1}{l|}{0.0276} & 0.1923 & 0.4053 & 0.2881 & 0.0497 \\
\multicolumn{1}{|c|}{} & \multicolumn{1}{l|}{} & \multicolumn{1}{l|}{FB-div} & 0.0788 & 0.6812 & 0.4570 & \multicolumn{1}{l|}{0.0810} & 0.1204 & 0.8570 & 0.6139 & \multicolumn{1}{l|}{0.1025} & 0.0471 & 0.7736 & 0.2884 & 0.0934 \\
\multicolumn{1}{|c|}{} & \multicolumn{1}{l|}{\multirow{2}{*}{hard}} & \multicolumn{1}{l|}{conventional} & 0.4117 & 0.4669 & 0.4446 & \multicolumn{1}{l|}{0.0116} & 0.5697 & 0.6512 & 0.6126 & \multicolumn{1}{l|}{0.0156} & 0.2387 & 0.4186 & 0.3161 & 0.0336 \\
\multicolumn{1}{|c|}{} & \multicolumn{1}{l|}{} & \multicolumn{1}{l|}{FB-div} & 0.1076 & 0.5571 & 0.4090 & \multicolumn{1}{l|}{0.0589} & 0.1641 & 0.7456 & 0.5598 & \multicolumn{1}{l|}{0.0768} & 0.1033 & 0.7381 & 0.3359 & 0.0932 \\ \hline
\multicolumn{1}{|c|}{\multirow{4}{*}{50}} & \multicolumn{1}{l|}{\multirow{2}{*}{easy}} & \multicolumn{1}{l|}{conventional} & 0.4176 & 0.5349 & 0.4624 & \multicolumn{1}{l|}{0.0287} & 0.5756 & 0.7061 & 0.6311 & \multicolumn{1}{l|}{0.0310} & 0.1834 & 0.3990 & 0.3036 & 0.0479 \\
\multicolumn{1}{|c|}{} & \multicolumn{1}{l|}{} & \multicolumn{1}{l|}{FB-div} & 0.2480 & 0.6456 & 0.4486 & \multicolumn{1}{l|}{0.0625} & 0.3565 & 0.8200 & 0.6086 & \multicolumn{1}{l|}{0.0720} & 0.1121 & 0.5538 & 0.3037 & 0.0749 \\
\multicolumn{1}{|c|}{} & \multicolumn{1}{l|}{\multirow{2}{*}{hard}} & \multicolumn{1}{l|}{conventional} & 0.4115 & 0.5185 & 0.4497 & \multicolumn{1}{l|}{0.0197} & 0.5638 & 0.6752 & 0.6067 & \multicolumn{1}{l|}{0.0202} & 0.2450 & 0.3852 & 0.3039 & 0.0329 \\
\multicolumn{1}{|c|}{} & \multicolumn{1}{l|}{} & \multicolumn{1}{l|}{FB-div} & 0.2662 & 0.6039 & 0.4401 & \multicolumn{1}{l|}{0.0506} & 0.3802 & 0.7756 & 0.5905 & \multicolumn{1}{l|}{0.0611} & 0.1504 & 0.6089 & 0.3174 & 0.0551 \\ \hline
\multicolumn{1}{|c|}{\multirow{4}{*}{100}} & \multicolumn{1}{l|}{\multirow{2}{*}{easy}} & \multicolumn{1}{l|}{conventional} & 0.3736 & 0.4355 & 0.4047 & \multicolumn{1}{l|}{0.0137} & 0.5225 & 0.5837 & 0.5540 & \multicolumn{1}{l|}{0.0150} & 0.2686 & 0.3905 & 0.3241 & 0.0270 \\
\multicolumn{1}{|c|}{} & \multicolumn{1}{l|}{} & \multicolumn{1}{l|}{FB-div} & 0.2789 & 0.5171 & 0.4028 & \multicolumn{1}{l|}{0.0312} & 0.4027 & 0.6834 & 0.5485 & \multicolumn{1}{l|}{0.0377} & 0.1909 & 0.4662 & 0.3301 & 0.0468 \\
\multicolumn{1}{|c|}{} & \multicolumn{1}{l|}{\multirow{2}{*}{hard}} & \multicolumn{1}{l|}{conventional} & 0.3954 & 0.4468 & 0.4153 & \multicolumn{1}{l|}{0.0096} & 0.5408 & 0.6027 & 0.5658 & \multicolumn{1}{l|}{0.0114} & 0.2971 & 0.3539 & 0.3241 & 0.0162 \\
\multicolumn{1}{|c|}{} & \multicolumn{1}{l|}{} & \multicolumn{1}{l|}{FB-div} & 0.3193 & 0.5110 & 0.4146 & \multicolumn{1}{l|}{0.0282} & 0.4436 & 0.6781 & 0.5628 & \multicolumn{1}{l|}{0.0348} & 0.2336 & 0.4193 & 0.3254 & 0.0263\\ \hline
\end{tabular}%
}
\caption{In this table, for each instance size, different hardness and algorithms, the minimum value, maximum value, average value and the standard deviation are included for three different feature values, which are the centroid\_mean\_distance\_to\_centroid, distance\_mean and nnd\_mean. The algorithms to be compared with are the conventional approach and the feature-based diversity maximization approach.}
\label{tb:range}
\end{table}

\begin{figure}
\centering
\includegraphics[height=4cm, width=8cm]{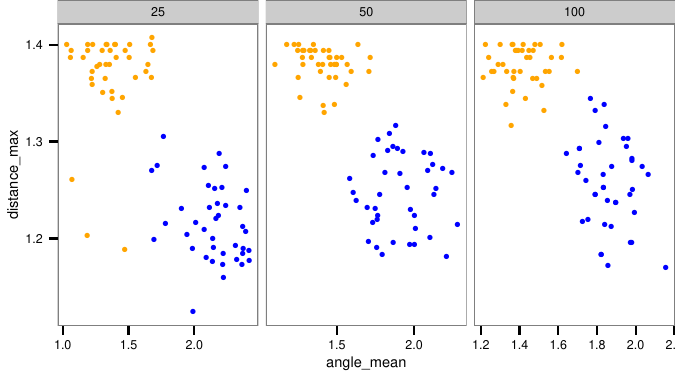}
\includegraphics[height=4cm, width=8cm]{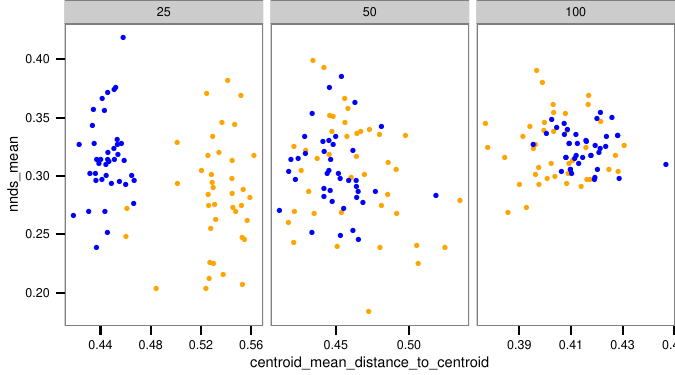}
\caption{$2$D Plots of feature combinations from the conventional approach. The blue dots and orange dots represent hard and easy instances respectively. The separation is mainly because the coverage of feature space is not large enough.}
\label{fig:2d_old}
\end{figure}

\begin{figure}
\centering
\includegraphics[height=4cm, width=8cm]{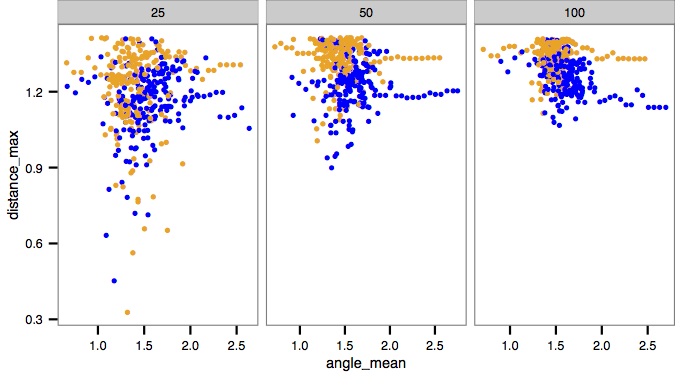}
\includegraphics[height=4cm, width=8cm]{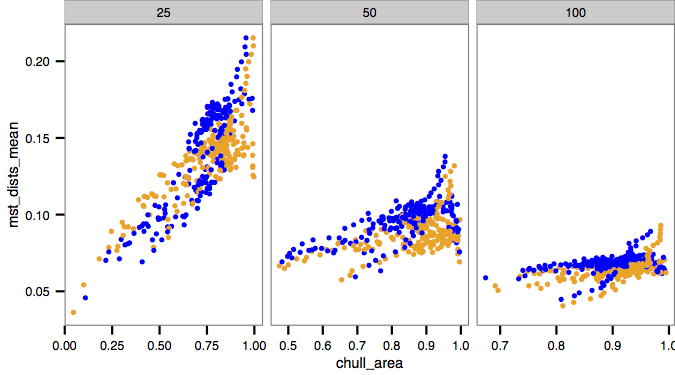}
\caption{$2$D Plots of feature combinations which provide a separation between easy and hard instances. The blue dots and orange dots represent hard and easy instances respectively.}
\label{fig:2d_good}
\end{figure}

\begin{figure}
\centering
\includegraphics[height=4cm, width=8cm]{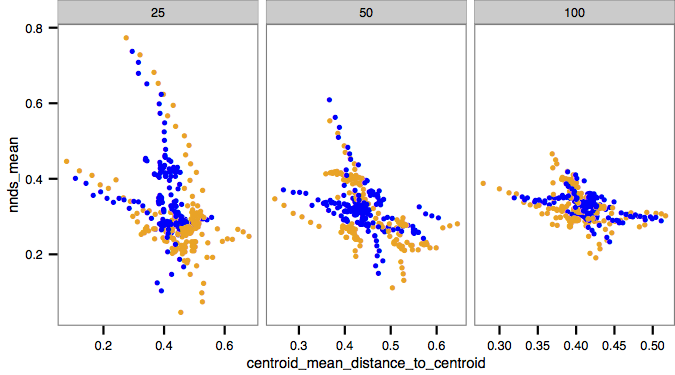}
\includegraphics[height=4cm, width=8cm]{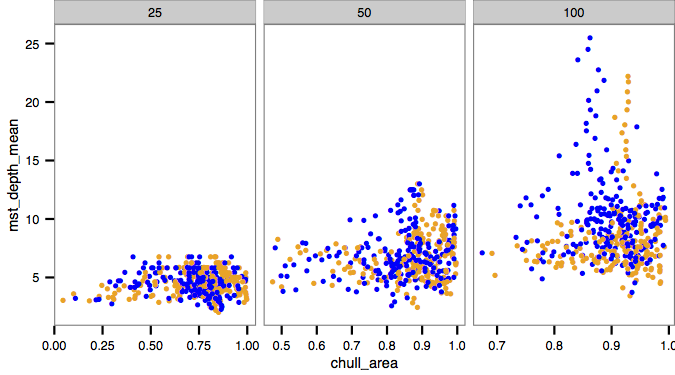}
\caption{$2$D plots of feature combinations which do not provide a clear separation between easy and hard instances. The blue dots and orange dots represent hard and easy instances respectively.}
\label{fig:2d_bad}
\end{figure}


In Table~\ref{tb:range}, some statistics are included to show the different ranges of the feature values of the instances found by the conventional approach and the diversity maximization EA. Since the conventional approach uses elitism to keep the best solution in each run, the data are gathered from $50$ independent runs of the conventional approach that generates TSP instance of acceptable quality for a certain problem size. The hard and easy instances here are classified with the same approximation ratio bound discussed previously in this section. The data to be compared with for the feature-based diversity optimization approach are obtained by a single run with population size $50$. In the table, three different features are selected from different categories to show the difference in range. From the table, it is clear that the diversity optimization approach generates TSP instances with feature values in a wider range compared to the conventional approach.

The above results suggest that the diversity optimization approach has resulted in a significant increase in the coverage over the feature space. Having the threshold for approximation ratios ($\alpha_e$ and $\alpha_h$) our method guarantees the hardness of the instances. These approximation thresholds are more extreme than the mean approximation values obtained by the conventional method. Furthermore, starting with initial population of duplicated instances and a hard coded threshold, the modified $(\mu+\lambda)$-EA is able to achieve hard instances with approximation ratio $1.347$, $1.493$ and $1.259$, respectively for instance size $25$, $50$ and $100$. The majority of the instances are clustered in a small region in the feature space while some other points are dispersed across the whole space. This is evident in the median values similar to the values for the instances with respect to conventional approach and with significantly larger range in feature value. The conventional approach has failed to explore certain regions in the feature space and missed some instances existing in those regions. Being able to discover all these instances spread in the whole feature space, our approach provides a strong basis for more effective feature based prediction.

As a result of the increased ranges and the similar gap in median feature values for hard and easy instances compared to the conventional instances, there is a strong overlap in the ranges of the features for easy and hard instances generated by the diversity optimization. This is observed in the results for \emph{mst\_dist\_mean} and \emph{cluster\_10pct\_distance\_to\_centroid} shown in Figure~\ref{fig:feature-range}.  Similar pattern holds for the other features as well. This prevents a good classification of problem instances based on single feature value. 


\section{Classification Based on Multiple Features}
\label{sec:exp}
As a single feature is not capable in clearly classifying the hard/easy instances, combinations of two or three different features are examined in the following.
Our analysis mainly focuses on combinations of the $7$ previously introduced features.

\subsection{Diversity Maximization over Single Feature Value}

Firstly, we represent the instances according to the combination of two different features in the $2$-dimensional feature value space (see Figure~\ref{fig:2d_good} for an example).

According to the observation and discussion in ~\citep{AMAI/Mersmann2013}, the two features \emph{distance \_max} and \emph{angle\_mean} can be considered together to provide an accurate classification of the hard and easy instances (see Figure~\ref{fig:2d_old}). Whereas after increasing the diversity over the seven different feature values and a wider coverage of the $2$D space is achieved, the separation of easy and hard instances is not so obvious. Comparing to Figure~\ref{fig:2d_good} and ~\ref{fig:2d_bad}, the clear separation shown in Figure~\ref{fig:2d_old} is mainly due to the instances generated from the conventional approach do not have a good coverage over the feature space. The clusters of dots representing hard and easy instances have some overlapping as shown in the left graphs of Figure~\ref{fig:2d_good}. There are large overlapping areas lying between the two groups of instances. Another example of some separation given by two-feature combination is \emph{mst\_dists\_mean} and \emph{chull\_area} which measure the mean distance of the minimum spanning tree and the area of the convex hull. However, as the number of cities in an instance increases, the overlapping area becomes larger. It is hard to do classification based on this.

After examining the $21$ different combinations of two features out of the seven features, we found out that some combinations of two features provide a fair separation between hard and easy instances after increasing the diversity over different feature values. As shown in Figure~\ref{fig:2d_good}, taking both \emph{mst\_dists\_mean} and \emph{chull\_area} features into consideration, some separations can be spotted between hard and easy instances. However, most combinations are not able to give a clear classification between hard and easy instances, for example in Figure~\ref{fig:2d_bad}, neither the combination of features \emph{nnds\_mean} and \emph{centroid\_mean\_distance\_to\_centroid} nor features \emph{mst\_depth\_mean} and \emph{chull\_area} shows clear classification between instances of different hardness. Moreover, along with the instance size increment, the overlapping area of the dots standing for hard and easy instances grows.

Since the majority of two-feature combinations are not capable of classifying easy and hard instances, the idea of combining three different features is put forward. As in the analysis of two-feature combination, the values of the three selected features are plotted in $3$D space.

\begin{figure}
\centering
\includegraphics[width=5.8cm]{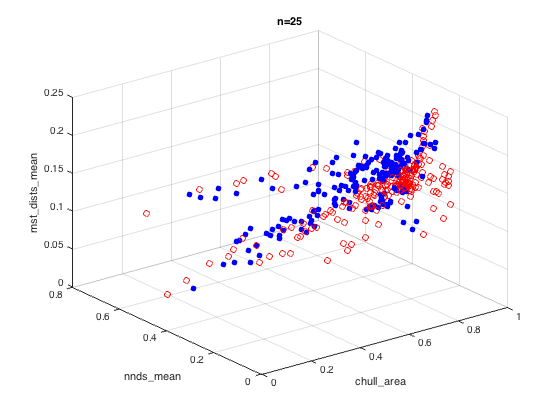}
\includegraphics[width=5.8cm]{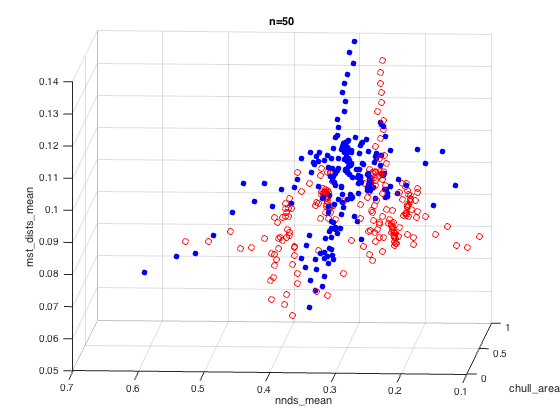}
\includegraphics[width=5.8cm]{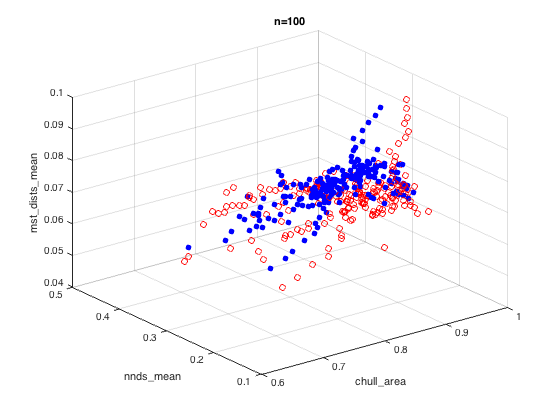}
\caption{$3$D Plots of combining experiment results from maximizing the diversity over features \emph{mst\_dists\_mean}, \emph{nnds\_mean} and \emph{chull\_area}, which provides a good separation of easy and hard instances. Hard and easy instances are represented as blue dots and red circles respectively. The SVM accuracy of SVM with RBF kernel is $0.9714$, $0.9857$ and $0.9667$.}
\label{fig:3d_good}
\end{figure}

\begin{figure}
\centering
\includegraphics[width=6cm]{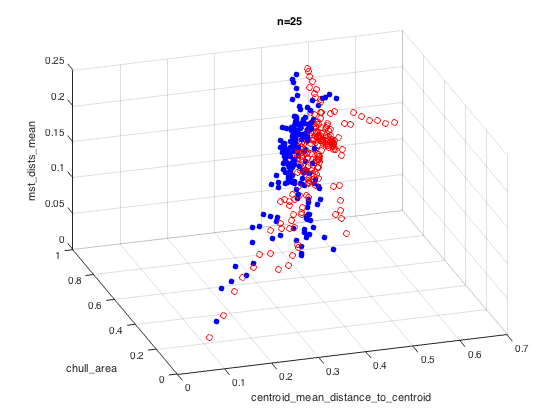}
\includegraphics[width=6cm]{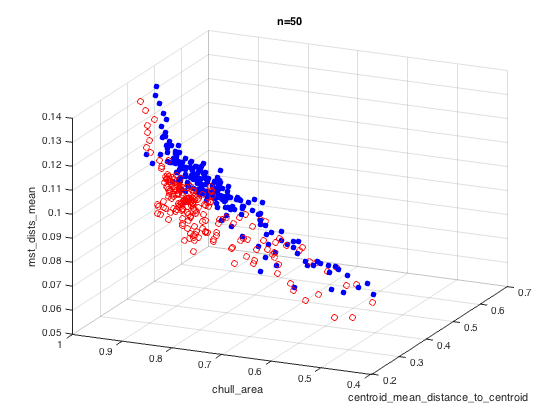}
\includegraphics[width=6cm]{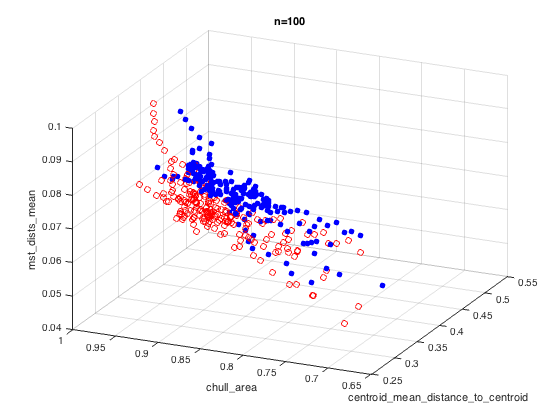}
\caption{$3$D Plots of combining experiment results from maximizing the diversity over features \emph{mst\_dists\_mean}, \emph{chull\_area} and \emph{centroid\_mean\_distance\_to\_centroid}, which provides a good separation of easy and hard instances. Legend is the same as that in Figure~\ref{fig:3d_good}. The SVM accuracy of SVM with RBF kernel is $0.9714$, $0.9714$ and $0.8381$.}
\label{fig:3d_good2}
\end{figure}

\begin{figure}
\centering
\includegraphics[width=6cm]{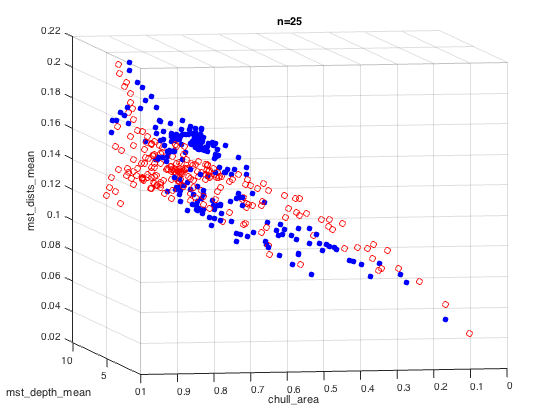}
\includegraphics[width=6cm]{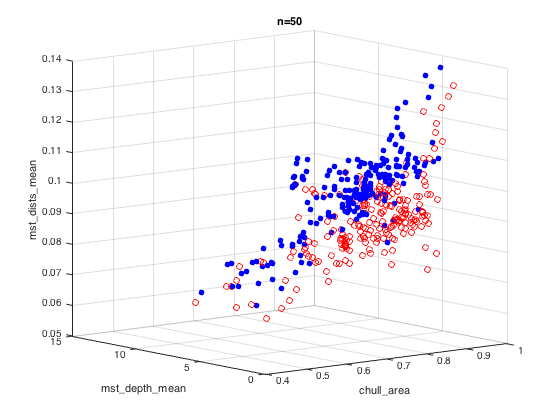}
\includegraphics[width=6cm]{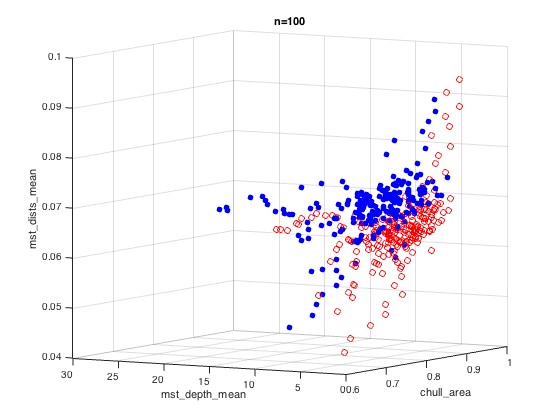}
\caption{$3$D Plots of combining experiment results from maximizing the diversity over features \emph{mst\_dists\_mean}, \emph{chull\_area} and \emph{mst\_depth\_mean}, which provides a good separation of easy and hard instances. Legend is the same as that in Figure~\ref{fig:3d_good}. The SVM accuracy of SVM with RBF kernel is $0.9476$, $0.9738$ and $0.9833$.}
\label{fig:2d-3d}
\end{figure}

By considering a third feature in the combination, in the $35$ different combinations, it is clear that there are some separations between the two groups of $210$ instances from the $3$D-plots. A good selection of features results in an accurate classification of the instances. The three-feature combinations with the features measuring statistics about minimum spanning tree always provide good separation between hard and easy instances as shown in Figure~\ref{fig:3d_good} and Figure~\ref{fig:3d_good2}. Although there is an overlapping in the area between the two clusters of hard and easy instances, from the $3$D-plots, we can spot some areas where there are only dots for instances of certain hardness.

Taken another feature value into consideration, the two-feature combination that is not able to provide good separation can give some clear classification in hard and easy instances. An example illustrating this is included as Figure~\ref{fig:2d-3d}, where together with an additional feature \emph{mst\_dists\_mean}, the two-feature combination of features \emph{mst\_depth\_mean} and \emph{chull\_area} shows a clear separation between easy and hard instances comparing to the results shown in the left graph in Figure~\ref{fig:2d_bad}.

From the investigation of both the two-feature combination and three-feature combination, we found out that the range of feature values for larger TSP instances is smaller. Some of the good combinations for classifying the hardness of smaller instances may not work for larger instances, such as centroid features which perform well when combining with another feature in classifying the hardness of instances of $25$ cities while do not show a clear separation with instance size $50$ and $100$ in our study. However, there exist some three-feature combinations that give good classification of easy and hard instances without regarding to the instance size, for example \emph{mst\_dists\_mean}, \emph{chull\_area} and \emph{nnds\_mean}, and \emph{mst\_dists\_mean}, \emph{chull\_area} and \emph{mst\_depth\_mean}.

\subsection{Diversity Maximization over Multiple Feature Values}
\label{sec:weighted}

In order to examine the relationship between feature combination and hardness of the instances, a weighted population diversity based on multiple features is introduced. The weighted population diversity for a certain set of features $\{f_1, f_2, ... ,f_k\}$ is defined as the weighted sum of the normalised population diversity over these $k$ features. The contribution of an instance $I$ to the weighted population diversity is defined as
$$d'(I,P)=\sum^{k}_{i=1}(w_i\cdot d_{f_i}(I,P)),$$
where $d_{f_i}(I,P)$ denotes the normalised contribution to the population diversity $d(I,P)$ over certain feature $i$ and $w_i$ represents the weight of feature $i$. The contribution of an individual to the population diversity on certain feature is normalised based on the maximum population diversity on the feature, in order to reduce the bias among different features.

This weighted population diversity is used in Algorithm~\ref{EA} to gain some insight of the relationship between features combination and instance quality. The same parent and offspring population sizes are used for these experiments, which are $\mu=30$ and $\lambda=5$. The instance sizes examined are still $25$, $50$ and $100$. The \topt algorithm is executed five times to obtain the approximation quality. In the experiments, \EAd execute for $10,000$ generation as previous. Since it is shown in Section~\ref{sec:range} that a combination of three features is able to provide a good separation between hard and easy instances, some of the good three-feature combinations are chosen for exploration. The weight distributions for $\{f_1,f_2,f_3\}$ considered in the experiments are $\{1,1,1\}$, $\{2,1,1\}$, $\{1,2,1\}$, $\{1,1,2\}$, $\{2,2,1\}$, $\{2,1,2\}$, $\{1,2,2\}$. The same hardness thresholds are used in these experiments as previous. After the seven independent runs for easy and hard instances, the final solution sets are put together. Therefore the results set has $210$ instances for each instance size and hardness type, which is the same as previous experiments. The results are plotted in $3$D space and compared to the previous experiments on single feature discussed in Section~\ref{sec:approach} and ~\ref{sec:range}.

The weighted population diversity offers a way to examine the overlapping area of hard and easy instances. With the weighting technique, it takes consideration about the relationship between the different features examined. Since most of these features are not independent from each others and the weighted population diversity considers multiple features at the same time, it is predictable that with the weighted population diversity the extreme value for each single feature may not reach.

\begin{figure}
\centering
\includegraphics[width=6cm]{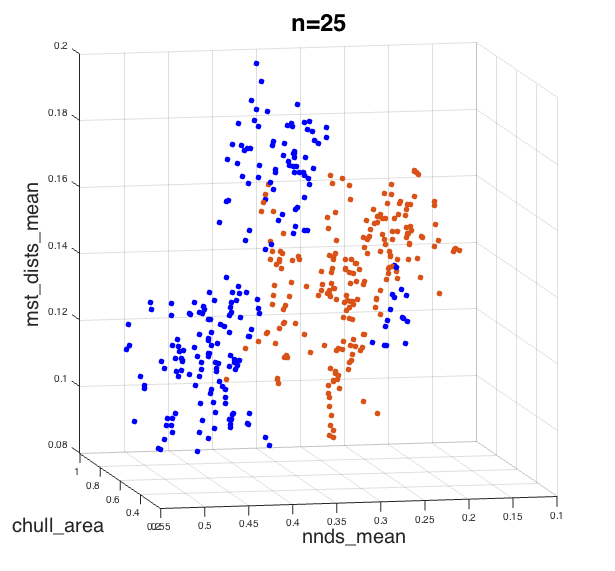}
\includegraphics[width=6cm]{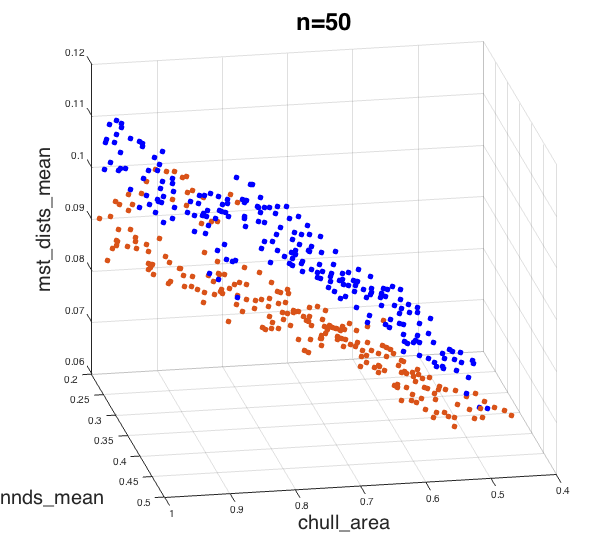}
\includegraphics[width=6cm]{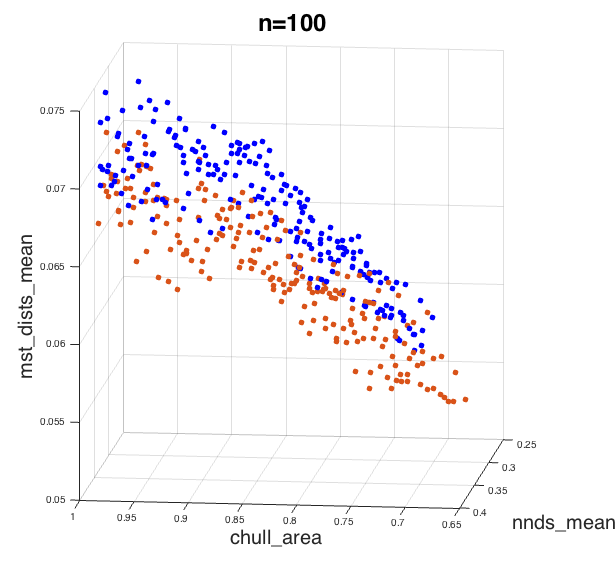}
\caption{$3$D Plots of combining experiment results from maximizing the diversity over features \emph{mst\_dists\_mean}, \emph{nnds\_mean} and \emph{chull\_area} with considering of weighting, which provides a good separation of easy and hard instances. Hard and easy instances are represented as blue dots and orange dots respectively.}

\label{fig:weighted}
\end{figure}

An example is shown in Figure~\ref{fig:weighted} focusing on maximizing the weighted population diversity over the combination of features \emph{mst\_dists\_mean}, \emph{nnds\_mean} and \emph{chull\_area}, which is shown to be a good combination for separating the hard and easy instances. From the comparison between Figure~\ref{fig:3d_good} and Figure~\ref{fig:weighted}, we can see that although the results from maximizing weighted population diversity does not cover a wider search space, it provides a detailed insight into the intersection between the hard and easy instances. The $3$D plots of different instance sizes show that the combination of these three certain features provide a clear separation between hard and easy instances. There are some overlapping areas in the search space, but it is clear that this combination of features provide some hints for predicting of hard or easy instances. 
It is expected that additional features might increase the separability. Because each additional feature contributes to an additional constraint on the hard/easy classification. However, due to visualization limitations we consider up to 3 features for graphical analysis.


\section{Instances Classification Using Support Vector Machine}
\label{sec:svm}

Support vector machines (SVMs) are well-known supervised learning models in machine learning which can be used for classification, regression and outliers detection ~\citep{Cortes95,Gunn98supportvector}. In order to quantify the separation between instances of different hardness based on the feature values, SVM models are constructed for each combination of features.

\subsection{Linear SVM}
The linear classifier is the first model tried in classifying the dataset. In SVM the linear classifiers that can separate the data with maximum margin is termed as the optimal separating hyper-plane. From the plots in Figure~\ref{fig:2d_good},~\ref{fig:2d_bad},~\ref{fig:3d_good} and~\ref{fig:3d_good2}, it is clear that none of the datasets are linearly separable. Taken the trade-off between maximizing the margin and minimizing the number of misclassified data points into consideration, the soft-margin SVM is used for classification. 

Let $ACC_n$ be the training accuracy of a feature combination in separating the hard and easy instances of size $n$. We define $ACC_n$ as the ratio of number of instances which are correctly classified by the model to the total number of instances in the dataset. All classification experiments are done in R with library$\{e1071\}$~\citep{Re1071}. 
The training data of the SVM models are the population of 420 instances generated as in Section~\ref{sec:approach} and the training accuracy is regarded as a quantified measurement of the separation between hard and easy instances.
The feature combinations used for classification are the $21$ two-feature combinations and $35$ three-feature combinations discussed in Section~\ref{sec:exp}.

From experiment results, $ACC_{25}$ for two-feature combinations lie in the range of $0.5095$ to $0.7548$ with an average accuracy of $0.6672$, while the $ACC_{25}$ for three-feature combinations lie between $0.6286$ to $0.7786$ with average value $0.7079$. In the case of instances with city number of $50$, two-feature combination results in $ACC_{50}$ lying in the range of $0.5286$ to $0.7738$ with an average of $0.6544$ while $ACC_{50}$ of three-feature combinations are from $0.5381$ to $0.85$ with average accuracy equal to $0.6969$. For larger instance size, $ACC_{100}$ are in the range between $0.5738$ and $0.8119$ with average $0.6986$ for two-feature combinations, whereas those for three-feature combinations lie in the scope of $0.6238$ to $0.8524$ with average $0.7382$.

Although three-feature combinations show better accuracy in separation of hard and easy instances than those two-feature combinations, there is no significant difference in $ACC$ for two-feature combinations and three-feature combinations. Moreover, the general low accuracy implies the high possibility that the linear models are not suitable for separating the hard and easy instances based on most of the feature combinations.

We then move to applying kernel function for non-linear mapping of the feature combination.

\subsection{Nonlinear Classification with RBF Kernel}
The linearly non-separable features can become linearly separable after mapped to a higher dimension feature space. The Radial Basis Function (RBF) kernel is one of the well-known kernel function used in SVM classification. 

There are two parameters need to be selected when applying RBF, which are $C$(cost) and $\gamma$. The parameter setting for RBF is crucial, since increasing $C$ and $\gamma$ leads to accurate separation of the training data but at the same time causes over-fitting~\cite{CHERKASSKY2004113}.
The SVMs here are generated for quantifying the separation rate between hard and easy instances rather than constructing a model for classification of other instances.
After some initial trials, $(C,\gamma)$ is set to $(100,2)$ in all the tests to avoid over-fitting to the selected training set.
This parameter setting may not be the best parameters for the certain feature combination in SVM classifying, but it helps us to gain some understanding of the separation of hard and easy instances generated from previous experiments based on the same condition. 

Table~\ref{tb:acc_2d} and ~\ref{tb:acc_3d} show the accuracy of different two features or three features combination in hard and easy instances separation. 
With RBF kernel, SVM with certain parameter setting can generate a model separating the dataset with average accuracy of $0.8170$, $0.8244$ and $0.8346$ in $2$D feature space for instance size $25$, $50$ and $100$ respectively. Whereas with three features, SVM with the same parameter setting provides a separation with average accuracy of $0.9503$, $0.9584$ and $0.9422$ for instance size $25$, $50$ and $100$ respectively.

From the results, it can be concluded that there are better separations between hard and easy instances in the $3$D feature space.


\begin{table*}
\centering
	\tiny
    \begin{tabular}{| l | l | c | c | c |}
    \hline
    Feature 1 & Feature 2 & $ACC_{25}$ & $ACC_{50}$ & $ACC_{100}$ \\ \hline
    angle\_mean & centroid\_mean\_distance\_to\_centroid & 0.8476 & 0.8071 & 0.8071 \\
    angle\_mean & chull\_area & 0.7857 & 0.7810 & 0.7929 \\
    angle\_mean & cluster\_10pct\_mean\_distance\_to\_centroid & 0.7810 & 0.7786 & 0.8000 \\
    angle\_mean & mst\_depth\_mean & 0.7524 & 0.7381 & 0.8000 \\
    angle\_mean & nnds\_mean & 0.8167 & 0.8833 & 0.8452 \\
    angle\_mean & mst\_dists\_mean & 0.8119 & 0.8024 & 0.8405 \\
    centroid\_mean\_distance\_to\_centroid & chull\_area & 0.8619 & 0.7667 & 0.8381 \\
    centroid\_mean\_distance\_to\_centroid & cluster\_10pct\_mean\_distance\_to\_centroid & 0.8524 & 0.8357 & 0.7548 \\
    centroid\_mean\_distance\_to\_centroid & mst\_depth\_mean & 0.8381 & 0.7643 & 0.8095 \\
    centroid\_mean\_distance\_to\_centroid & nnds\_mean & 0.8786 & 0.9524 & 0.8476 \\
    centroid\_mean\_distance\_to\_centroid & mst\_dists\_mean & 0.8905 & 0.8571 & 0.8762 \\
    chull\_area & cluster\_10pct\_mean\_distance\_to\_centroid & 0.8000 & 0.7881 & 0.8548 \\
    chull\_area & mst\_depth\_mean & 0.7429 & 0.7429 & 0.7571 \\
    chull\_area & nnds\_mean & 0.8071 & 0.8905 & 0.8452 \\
    chull\_area & mst\_dists\_mean & 0.8619 & 0.8643 & 0.9024 \\
    cluster\_10pct\_mean\_distance\_to\_centroid & mst\_depth\_mean & 0.7619 & 0.7714 & 0.7929 \\
    cluster\_10pct\_mean\_distance\_to\_centroid & nnds\_mean & 0.8190 & 0.8833 & 0.8643 \\
    cluster\_10pct\_mean\_distance\_to\_centroid & mst\_dists\_mean & 0.8095 & 0.8095 & 0.8738 \\
    mst\_depth\_mean & nnds\_mean & 0.7786 & 0.8595 & 0.8405 \\
    mst\_depth\_mean & mst\_dists\_mean & 0.8095 & 0.8214 & 0.8810 \\
    nnds\_mean & mst\_dists\_mean & 0.8500 & 0.9143 & 0.9024 \\
\hline
 \end{tabular}
       \caption[Table caption text]{This table lists the accuracy of SVM with RBF kernel separating the hard and easy instances in 21 different two-feature space. }
    \label{tb:acc_2d}
\end{table*}

\begin{table*}
\centering
	\tiny
    \begin{tabular}{| l | l | l | c | c | c |}
    \hline
    Feature 1 & Feature 2 & Feature 3 & $ACC_{25}$ & $ACC_{50}$ & $ACC_{100}$ \\ \hline
    angle\_mean & centroid\_mean\_distance\_to\_centroid & chull\_area & 0.9500 & 0.9190 & 0.9452 \\
    angle\_mean & centroid\_mean\_distance\_to\_centroid & cluster& 0.9405 & 0.9357 & 0.8214 \\
    angle\_mean & centroid\_mean\_distance\_to\_centroid & mst\_depth\_mean & 0.9548 & 0.9548 & 0.9214 \\
    angle\_mean & centroid\_mean\_distance\_to\_centroid & nnds\_mean & 0.9452 & 0.9952 & 0.9833 \\
    angle\_mean & centroid\_mean\_distance\_to\_centroid & mst\_dists\_mean & 0.9571 & 0.9500 & 0.9524 \\
    angle\_mean & chull\_area & cluster & 0.9524 & 0.9310 & 0.8881 \\
    angle\_mean & chull\_area & mst\_depth\_mean & 0.9357 & 0.9238 & 0.9500 \\
    angle\_mean & chull\_area & nnds\_mean & 0.9405 & 0.9714 & 0.9571 \\
    angle\_mean & chull\_area & mst\_dists\_mean & 0.9667 & 0.9619 & 0.9143 \\    
    angle\_mean & cluster& mst\_depth\_mean & 0.9214 & 0.9143 & 0.9810 \\
    angle\_mean & cluster& nnds\_mean & 0.9476 & 0.9690 & 0.9333 \\
    angle\_mean & cluster& mst\_dists\_mean & 0.9571 & 0.9143 & 0.9405 \\
    angle\_mean & mst\_depth\_mean & nnds\_mean & 0.9310 & 0.9762 & 0.9238 \\
    angle\_mean & mst\_depth\_mean & mst\_dists\_mean & 0.9476 & 0.9262 & 0.9476 \\
    angle\_mean & nnds\_mean & mst\_dists\_mean & 0.9429 & 0.9762 & 0.8833 \\
    centroid\_mean\_distance\_to\_centroid & chull\_area & cluster& 0.9476 & 0.9333 & 0.9310 \\
    centroid\_mean\_distance\_to\_centroid & chull\_area & mst\_depth\_mean & 0.9595 & 0.8762 & 0.9762 \\
    centroid\_mean\_distance\_to\_centroid & chull\_area & nnds\_mean & 0.9667 & 0.9881 & 0.9929 \\
    centroid\_mean\_distance\_to\_centroid & chull\_area & mst\_dists\_mean & 0.9714 & 0.9714 & 0.8381 \\    
    centroid\_mean\_distance\_to\_centroid & cluster& mst\_depth\_mean & 0.9476 & 0.9286 & 0.8571 \\
    centroid\_mean\_distance\_to\_centroid & cluster& nnds\_mean & 0.9643 & 0.9905 & 0.8810 \\
    centroid\_mean\_distance\_to\_centroid & cluster& mst\_dists\_mean & 0.9500 & 0.9595 & 0.9190 \\    
    centroid\_mean\_distance\_to\_centroid & mst\_depth\_mean & nnds\_mean & 0.9500 & 0.9881 & 0.9595 \\
    centroid\_mean\_distance\_to\_centroid & mst\_depth\_mean & mst\_dists\_mean & 0.9548 & 0.9548 & 0.9595 \\
    centroid\_mean\_distance\_to\_centroid & nnds\_mean & mst\_dists\_mean & 0.9667 & 1.0000 & 0.9952 \\
    chull\_area & cluster& mst\_depth\_mean & 0.9286 & 0.9524 & 0.9333 \\
    chull\_area & cluster& nnds\_mean & 0.9524 & 0.9667 & 0.9667 \\
    chull\_area & cluster& mst\_dists\_mean & 0.9595 & 0.9595 & 0.9929 \\    
    chull\_area & mst\_depth\_mean & nnds\_mean & 0.9381 & 0.9857 & 0.9476 \\
    chull\_area & mst\_depth\_mean & mst\_dists\_mean & 0.9476 & 0.9738 & 0.9833 \\
    chull\_area & nnds\_mean & mst\_dists\_mean & 0.9714 & 0.9857 & 0.9667 \\
    cluster & mst\_depth\_mean & nnds\_mean & 0.9214 & 0.9857 & 0.9738 \\
    cluster & mst\_depth\_mean & mst\_dists\_mean & 0.9500 & 0.9476 & 0.9643 \\
    cluster & nnds\_mean & mst\_dists\_mean & 0.9643 & 0.9833 & 0.9976 \\
    mst\_depth\_mean & nnds\_mean & mst\_dists\_mean & 0.9429 & 0.9929 & 0.9929 \\
\hline
 \end{tabular}
       \caption[Table caption text]{This table lists the accuracy of SVM with RBF kernel separating the hard and easy instances in 35 different three-feature space. The cluster feature in the table refers to the cluster\_10pct\_mean\_distance\_to\_centroid feature.}
    \label{tb:acc_3d}
\end{table*}

\section{Diversity optimization for Instance Hardness}
\label{sec:divAP}

In the experiments in previous sections, the main focus of diversity maximization is the feature values. The quality of solutions is guaranteed by a predefined threshold. In order to have an insight into the relationship between the feature values and problem hardness, we conduct another experiment about population diversity optimization with instance hardness. The experiment is based on the TSP and Algorithm~\ref{EA} as well. In this case, the approximation ratio is taken as a feature value. Maximizing the diversity over approximation ratio results in a set of individuals with different quality. By doing this, we obtain knowledge about the relationship between feature values and instance hardness from another point of view.

\begin{algorithm}[t]
{
   	Initialize the population $P$ with $\mu$ TSP instances of certain size.\\
	Let $C \subseteq P$ where $|C| = \lambda$.\\
	For each $ I \in C$, produce an offspring $I'$ of $I$ by mutation and add $I'$ to $P$. \\
	While $|P| > \mu$, remove an individual $I=\arg \min_{J \in P} d_{ar}(J,P)$ uniformly at random.\\ 	
 	Repeat step 2 to 4 until termination criterion is reached.\\
} 
 \caption{$(\mu+\lambda)$-$EA_{DA}$}
\label{alg:EA-divApp}
\end{algorithm}

In this case, the algorithm needs to be modified in the solution quality check phase. The evolutionary algorithm without hard coded solution quality threshold is shown in Algorithm~\ref{alg:EA-divApp}. The population diversity measurement $d_{ar}(I,P)$ in step 4 follows the formulation proposed in Section~\ref{sec:approach} with the approximation ratio as the target feature value. In the survivor selection phase, the individual that contributes the least to the population diversity over approximation ratio is removed.

For the purpose of reasonable coverage over the whole space, the population size and offspring population size is set to $400$ and $10$ respectively, in this experiment. The other parameters are all kept the same with previous experiments. Then the R program is run for 30,000 generations to obtain some stable results.

\begin{figure*}
\centering
\includegraphics[width=6cm]{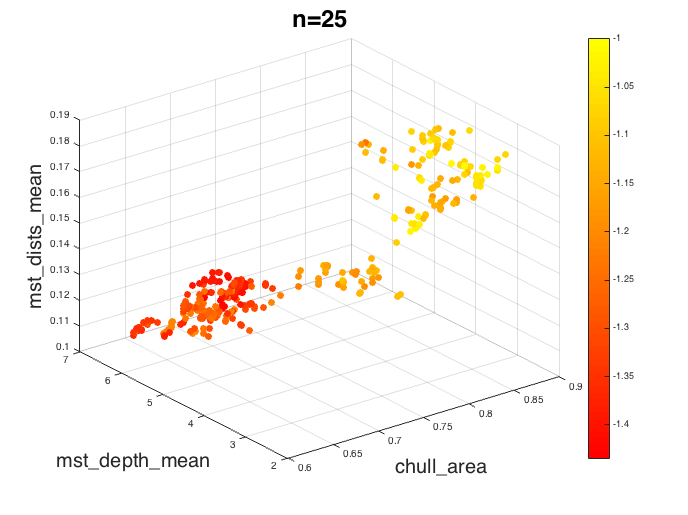}
\includegraphics[width=6cm]{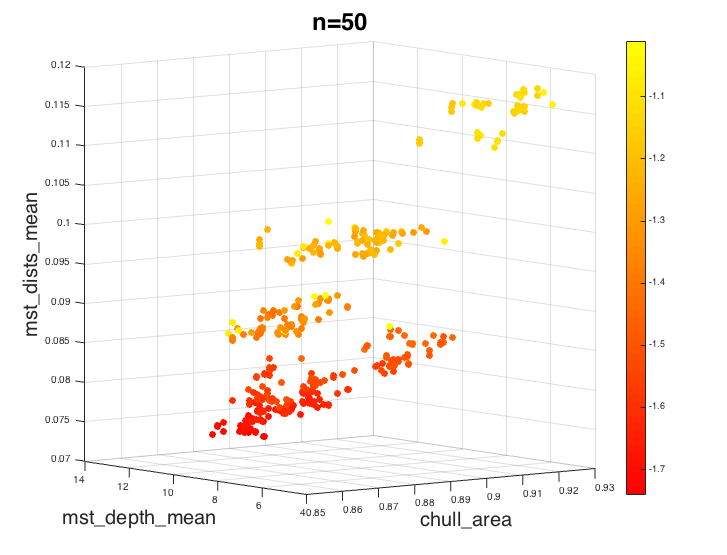}
\includegraphics[width=6cm]{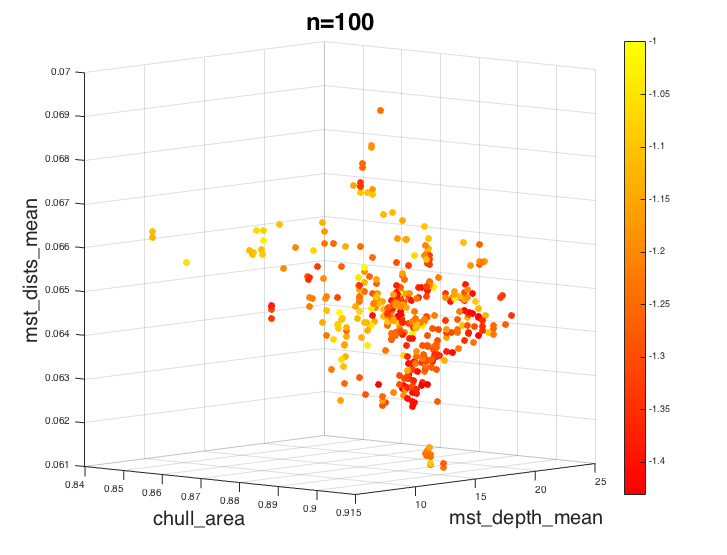}

\caption{$3$D Plots of experiment results from maximizing the diversity over approximation ratio in the feature space of feature combination \emph{chull\_area}, \emph{mst\_dist\_mean} and \emph{mst\_depth\_mean}. The color of the dot reflects the hardness of the problem and the easier problem is indicated with lighter color.}
\label{fig:divApp}
\end{figure*}

\begin{figure*}
\centering
\includegraphics[width=6cm]{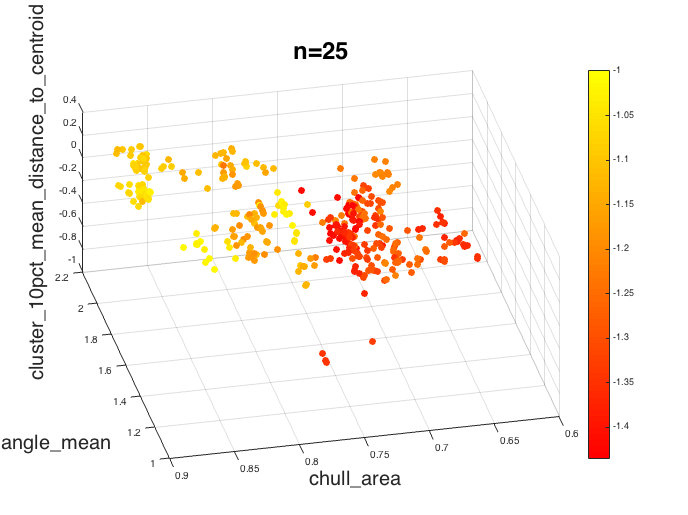}
\includegraphics[width=6cm]{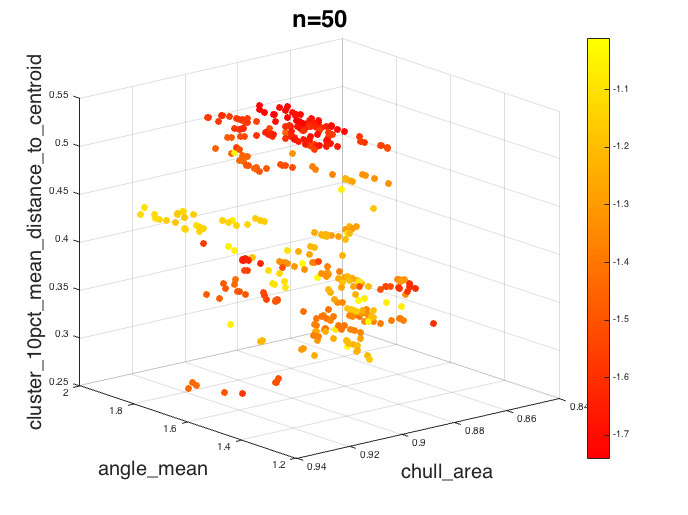}
\includegraphics[width=6cm]{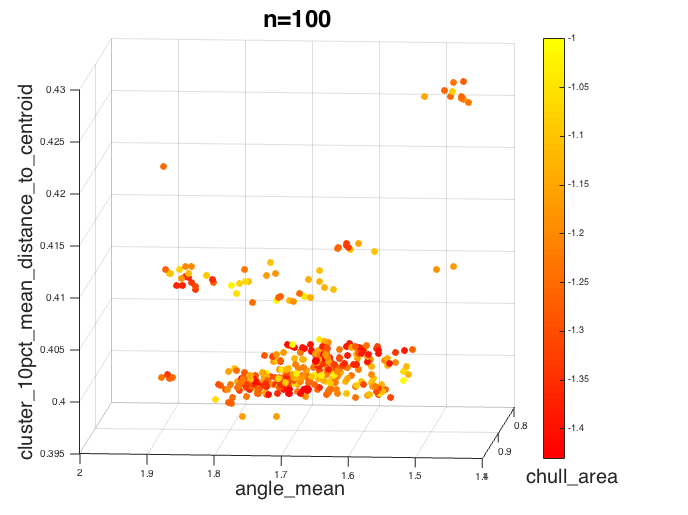}
\caption{$3$D Plots of experiment results from maximizing the diversity over approximation ratio in the feature space of feature combination \emph{chull\_area}, \emph{angle\_mean} and \emph{cluster\_$10$pct\_mean\_distance\_to\_centroid}. The colors of the dots reflect the hardness of the problem.}
\label{fig:divApp-2}
\end{figure*}

Figure~\ref{fig:divApp} and~\ref{fig:divApp-2} show some example $3$D plots of the experimental results. The $400$ individuals are plotted in the $3$D feature space with consideration over different feature combinations. The vertical colorbar lying on the right side of each plot illustrates the relationship between color and problem hardness and indicates the mapping of approximation ratio into the different colors. The dots in lighter color imply easier instances. 

The plots in Figure~\ref{fig:divApp} present the resulting instances in the space of feature \emph{chull\_area}, \emph{mst\_dist\_mean} and \emph{mst\_depth\_mean}. There is clear separation between the red dots and yellow dots in the figure for different problem sizes. The red and yellow dots refer to instances with extreme approximation ratio in each case. For larger instances, there is no clear cluster of instances with different hardness but we can still find areas where hard or easy instances gather. On the other hand, for problem instances with approximation ratio lying in the range of $1.1$ to $1.2$, $1.2$ to $1.4$ and $1.15$ to $1.2$ which are represented by orange dots in the plots, the locations are hard to classify.

The plots in Figure~\ref{fig:divApp-2} show the relationship of problem hardness with the feature combination \emph{chull\_area}, \emph{angle\_mean} and \emph{cluster\_$10$pct\_mean\_distance\_to\_centroid}. Similar observation can be obtained. The separation for problem size $25$ is clear. However, when the number of cities in each instance increases, the overlapping area becomes larger.

Algorithm~\ref{alg:EA-divApp} allows us to generate a set of solutions with different hardness. The approximation ratio for problem size $25$, $50$ and $100$ is maximized to $1.434$, $1.739$ and $1.430$ in this experiment which implies the populations cover a even wider range than those from previous experiments.

\section{Conclusions}
\label{sec:conclusions}
With this paper, we have introduced a new methodology of evolving easy/hard instances which are diverse with respect to feature sets of the optimization problem at hand. 
Using our diversity optimization approach we have shown that the easy and hard instances obtained by our approach cover a much wider range in the feature space than previous methods. The diversity optimization approach provides instances which are diverse with respect to the investigated features. The proposed population diversity measurements provide a good evaluation of the diversity over single or multiple feature values.
Our experimental investigations for \topt and TSP provide evidence that the large set of diverse instances can be classified quite well into easy and hard instances when considering a suitable combination of multiple features which provide some guidance for predication as the next step. In particular, the SVM classification model built with the diverse instances that can classify TSP instances based on problem hardness provides a strong basis for future performance prediction models that lead to automatic algorithm selection and configuration. Building such models would require further experimentation to determine the minimal set of strong features that can predict performance accurately.

\section*{Acknowledgements}
This research has been supported by ARC grants DP140103400 and DP190103894.

\bibliographystyle{apalike}
\bibliography{main}

\end{document}